\documentclass{article}
\usepackage{spconf,amsmath,epsfig}

\usepackage{graphicx}
\usepackage{amsmath,amssymb} 
\usepackage{color}

\usepackage{epsfig}
\usepackage{bm}
\usepackage{stackrel}
\usepackage{textcomp}
\usepackage{color, colortbl}
\definecolor{LightCyan}{rgb}{0.88,1,1}
\usepackage{enumitem}
\newcommand{\norm}[1]{\left\lVert#1\right\rVert}

\begin{document}\sloppy

\def\x{{\mathbf x}}
\def\L{{\cal L}}

\title{Human-Centered Emotion Recognition in Animated GIFs}
%
\name{Zhengyuan Yang, Yixuan Zhang, Jiebo Luo}
\address{Department of Computer Science,
University of Rochester, Rochester NY 14627, USA\\
\{zyang39, jluo\}@cs.rochester.edu, yzh215@ur.rochester.edu}

\maketitle

%
\begin{abstract}
As an intuitive way of expression emotion, the animated Graphical Interchange  Format  (GIF) images have been widely used on social media. Most previous studies on automated GIF emotion recognition fail to effectively utilize GIF's unique properties, and this potentially limits the recognition performance. 
In this study, we demonstrate the importance of human related information in GIFs and conduct human-centered GIF emotion recognition with a proposed Keypoint Attended Visual Attention Network (KAVAN). The framework consists of a facial attention module and a hierarchical segment temporal module. 
The facial attention module exploits the strong relationship between GIF contents and human characters, and extracts frame-level visual feature with a focus on human faces. The Hierarchical Segment LSTM (HS-LSTM) module is then proposed to better learn global GIF representations. Our proposed framework outperforms the state-of-the-art on the MIT GIFGIF dataset. 
Furthermore, the facial attention module provides reliable facial region mask predictions, which improves the model's interpretability.
\end{abstract}
%
\section{Introduction}
The animated Graphical Interchange Format (GIF) images have been widely used on social media for online chatting and emotion expression~\cite{bakhshi2016fast,chen2017gifgif+}. The GIFs are short image sequences and are more light weighted compared to videos. Because of this, it can be used on social media with a lower time lag and required bandwidth. On the other hand, GIFs have a better ability to express emotions compared to still images because of the contained temporal information.
By analyzing over 3.9 million posts on Tumblr, Bakhshi~{\it et al.}~\cite{bakhshi2016fast} show that GIFs are significantly more engaging than other online media types. Because of GIF's popularity, many previous studies explire automated GIF emotion recognition. Most studies~\cite{jou2014predicting, chen2016predicting} extract visual representations for emotion recognition with pre-defined features or convolutional neural networks. Although previous approaches provide feasible solutions for GIF emotion recognition, they process GIFs as general videos and fail to utilize GIF's unique properties. We show this potentially limit the recognition performance and propose the human-centered GIF emotion recognition.

Human and human-like characters play an importance role in GIFs. A sampling on a GIF search engine GIPHY\footnote{https://giphy.com/} shows that a majority of GIFs contain clear human or cartoon faces. A previous study~\cite{tang2013deep} also reveals the importance of human faces in expressing emotions. Motivated by this, we explore human-centered GIF emotion recognition and improve recognition performance by focusing on informative facial regions. To be specific, we design a side task of facial region prediction in the proposed facial attention module, where estimated facial keypoints are used to represent human information and are fused with frame-level visual features.
\begin{figure}[t]
\begin{center}
   \centerline{\includegraphics[width=7.5cm]{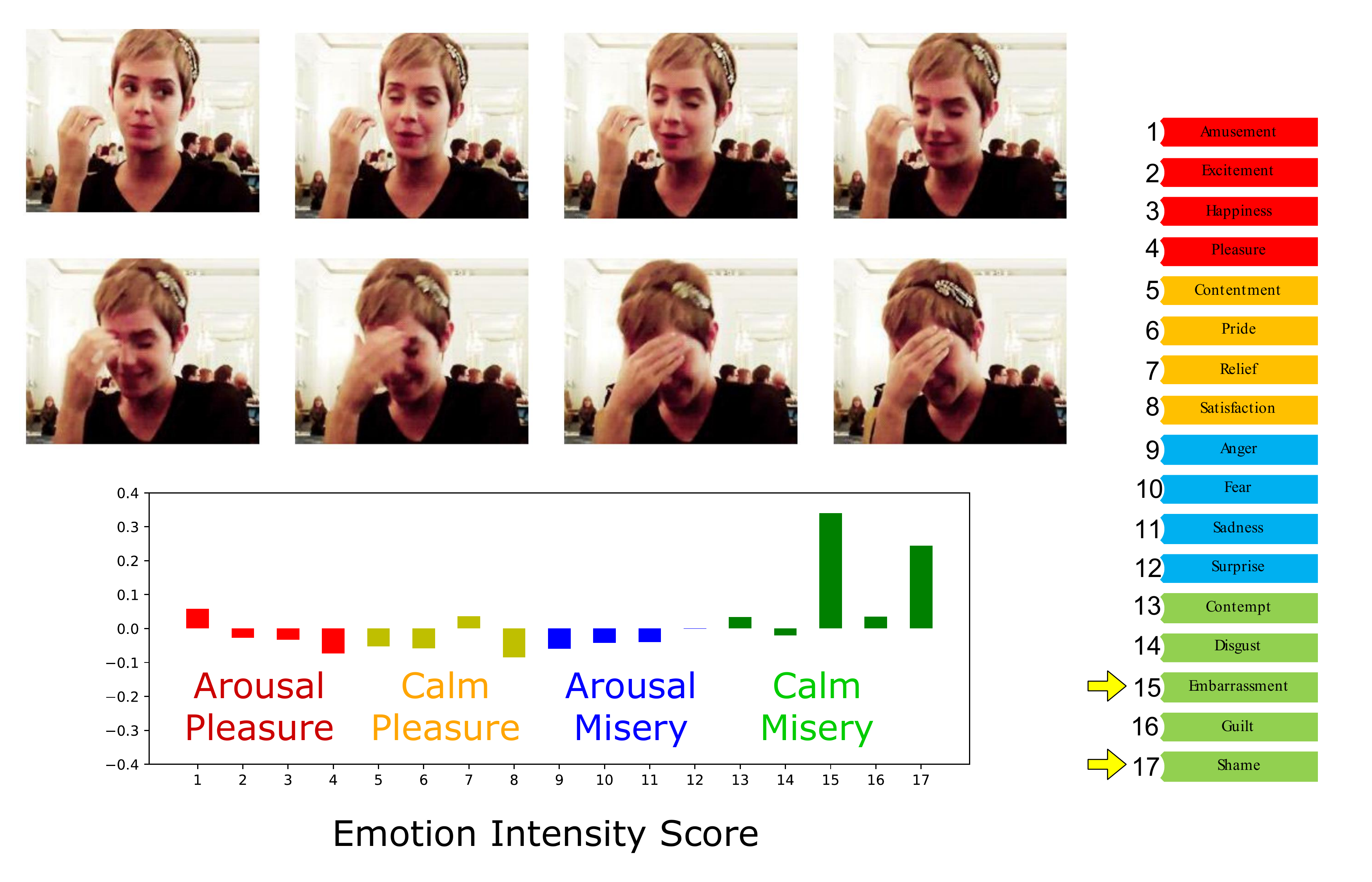}}
\end{center}
\vspace{-0.4in}
    \caption{We formulate GIF emotion recognition both as a classification task for coarse category prediction, and a regression task for emotion intensity score estimation. As shown in the bar chart, the four coarse GIF categories are represented by four different bar colors, and each bar shows the intensity score for one of the 17 annotated emotions.}
\vspace{-0.1in}
\label{fig:demo_gif}
\end{figure}
Combining human keypoints with appearance features has shown its effectiveness in related video analysis tasks~\cite{jhuang2013towards, cheron2015p}. A majority of methods merge keypoints as an extra input modality, and thus require keypoints to be complete and accurate. However, the quality of keypoints often can not be guaranteed, especially when keypoints are machine estimated instead of manually labeled. In the facial attention module, we propose to take estimated facial keypoints as the supervision for a facial region prediction side task, and use predicted regions as attention weights to further refine extracted frame-level visual features. 
As discussed in Section \ref{sec:kavan}, the soft attention fusion is naturally robust against keypoint incompleteness. We further include the keypoint estimation confidence scores in the heatmap generation stage, and make KAVAN robust with respect to inaccurate keypoints. In short, the facial regions predicted by the side task refine the visual features by assigning higher weights to informative facial regions. Furthermore, the predicted facial regions improve the method's interpretability by reliably localizing facial regions.

Another unique property for GIFs is its temporal conciseness. Unlike videos that contain a portion of `background frames' to better depict a complete story, GIFs are more compact and contain few `redundant frames'.
For example, emotions `embarrassment' and `shame' can only be correctly interpreted when jointly looking at all frames presented in Fig.~\ref{fig:demo_gif}. To better capture the temporal information from different segments of a GIF, we propose a Hierarchical Segment LSTM (HS-LSTM) structure as KAVAN's temporal module. GIFs are first evenly split into several temporal segments. The coarse local segment representation is then captured by HS-LSTM nodes. Finally a global GIF representation is learned with segment features from coarse- to fine-grained.

In this study, we propose the Keypoint Attended Visual Attention Network (KAVAN), which improves GIF emotion recognition performance by effectively utilizing GIF's unique properties. In the facial attention module, we utilize human information by merging estimated facial keypoints. 
Furthermore, we show that replacing the traditional LSTM layers in KAVAN with the proposed HS-LSTM structure can help better modeling temporal evolution in GIFs and further improve the recognition accuracy. Extensive experiments on the GIFGIF dataset prove the effectiveness of our methods.
\section{Related Work}
{\bf GIF Analysis.} Bakhshi~{\it et al.}~\cite{bakhshi2016fast} show that animated GIFs are more engaging than other social media types by studying over 3.9 million posts on Tumblr. Gygli~{\it et al.}~\cite{gygli2016video2gif} propose to automatically generate animated GIFs from videos with 100K user-generated GIFs and the corresponding video sources. The MIT's GIFGIF platform is frequently used for GIF emotion recognition studies. Jou~{\it et al.}~\cite{jou2014predicting} recognize GIF emotions using color histograms, facial expressions, image based aesthetics and visual sentiment. Chen~{\it et al.}~\cite{chen2016predicting} adopt 3D ConvNets to further improve the performance. The GIFGIF+ dataset~\cite{chen2017gifgif+} is a larger GIF emotion recognition dataset. At the time of this study, GIFGIF+ is not released.

\noindent{\bf Emotion Recognition.} 
Emotion recognition~\cite{zhao2014exploring,zhao2017continuous} has been an interesting topic for decades. On a large scale dataset~\cite{you2016building}, Rao~{\it et al.}~\cite{rao2016learning} propose a multi-level deep representations for emotion recognition. Multi-modal feature fusion~\cite{zhao2017learning} is also proved to be effective. 
Instead of modeling emotion recognition as a classification task~\cite{rao2016learning,you2016building}, Zhao~{\it et al.}~\cite{zhao2017learning} propose to learn emotion distributions instead, which alleviates the perception uncertainty problem that different people under different context may perceive different emotions from the same content. Regressing emotion intensity scores~\cite{jou2014predicting} is another effective approach. Han~{\it et al.}~\cite{han2017hard} propose a soft prediction framework for the perception uncertainty problem.
\section{Methodology}
In this section, we introduce the proposed Keypoint Attended Visual Attention Network (KAVAN), which consists of a facial soft attention module and a temporal module. For clarity, the soft attention module is first introduced with a traditional LSTM temporal module in Section \ref{sec:kavan}. We then introduce the novel temporal module in KAVAN, namely the Hierarchical Segment LSTM (HS-LSTM) in Section \ref{sec:hslstm}. Finally, we discuss the training objective and the refined problem setting for GIF emotion recognition.
\begin{figure}[t]
\begin{center}
   \centerline{\includegraphics[width=9cm]{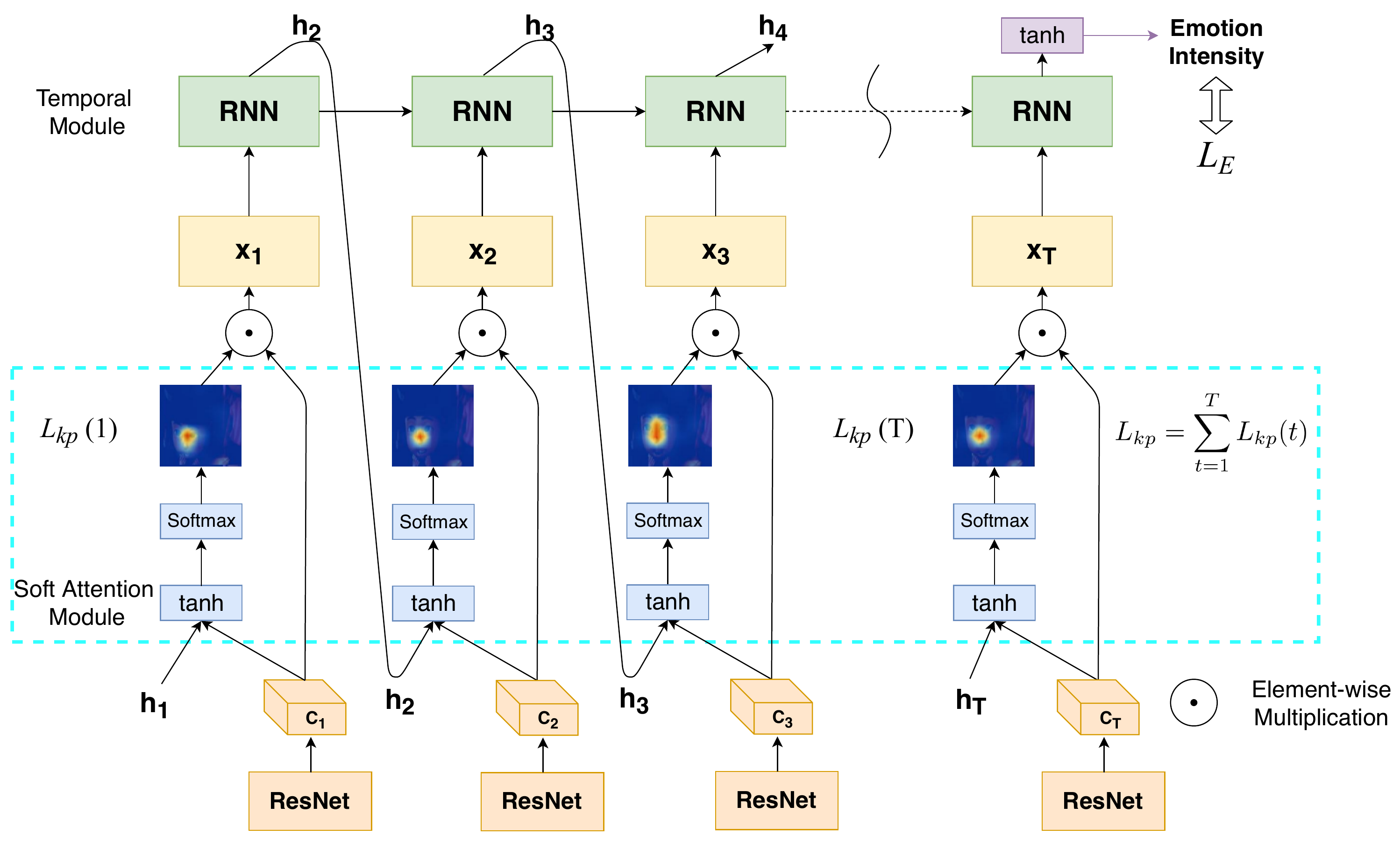}}
\end{center}
\vspace{-0.3in}
	\caption{The structure of the Keypoint Attended Visual Attention Network (KAVAN). Human centered visual feature $X_i$ is first obtained with the soft attention module draw in blue. The RNN temporal module consists of either a single LSTM layer or the proposed HS-LSTM then learns a global GIF representation for emotion recognition.}
\vspace{-0.2in}               
\label{fig:att_structure}
\end{figure}
\vspace{-9pt}
\subsection{Keypoint Attended Visual Attention Network}
\label{sec:kavan}
One unique property for GIFs is the frequent appearance of human and cartoon faces. More than $50\%$ of the GIFs in the MIT GIFGIF dataset contain human faces. Moreover, many in the remaining portion contain cartoon or personated animal characters that also have abundant facial expressions. Previous studies~\cite{bakhshi2016fast,bakhshi2014faces} also show a strong relationship between faces and the engagement level of social media contents. Motivated the importance of human faces in GIFs, we explore human-centered GIF emotion recognition. 

We represent human information as estimated facial keypoints, and propose a facial soft attention module in the Keypoint Attended Visual Attention Network (KAVAN) to utilize the information by fusing keypoints with extracted frame-level visual features. A number of video action recognition studies~\cite{jhuang2013towards,cheron2015p} have explored the fusion of keypoints and appearance features. However, previous studies require manually labeled accurate keypoints and might collapse with noisy estimated keypoints. The major challenge is that estimated keypoints can be {\it inaccurate} and {\it incomplete}, i.e. certain estimates could be wrong or missing because of occlusions or algorithm failures. In order to solve this challenge, the soft attention module in KAVAN is proposed to fuse the two modalities with attention mechanism. We first introduce the side task of facial region prediction. The predicted facial masks are then processed as attention masks to refine visual features. The soft attention module helps focusing on informative facial regions and thus contributes to GIF emotion recognition. 

The proposed KAVAN structure is shown in Fig.~\ref{fig:att_structure}. Following Temporal Segments Network (TSN)~\cite{wang2016temporal}, GIFs are first evenly split into $T$ segments and one frame is randomly sampled from each segments as network inputs. At each time-stamp $t$, a visual feature block $C_t^{H*W*D}$ is extracted with the backbone network~\cite{he2016deep}, and a facial region mask $\alpha_t$ is predicted. Extracted visual features are then refined by facial region mask $\alpha_t$ and fed into a temporal module for GIF emotion recognition. The temporal module can be as simple as a single LSTM layer, or other more effective structures as introduced in Section~\ref{sec:hslstm}. For clarity, we first introduce the base KAVAN structure with a single LSTM layer:
\begin{equation}
\left(
  \begin{matrix}
  i_t\\f_t\\o_t\\g_t
  \end{matrix}
\right)=
\left(
  \begin{matrix}
  \sigma\\\sigma\\\sigma\\\tanh
  \end{matrix}
\right)T_{d+D,4d}
\left(
  \begin{matrix}
  h_{t-1}\\x_t
  \end{matrix}
\right)
\end{equation}
\begin{equation}
c_t=f_t\odot c_{t-1}+i_t\odot g_t
\end{equation}
\begin{equation}
h_t=o_t\odot \tanh(c_t)
\end{equation}
\begin{equation}
x_t = \sum_{k=1}^{H*W} \left(\alpha_t(k)+w_{res}\right)\ C_t(k)
\end{equation}
where $i_t,f_t,o_t,c_t,h_t$ are the input, forget, output, memory and hidden states. $D$ is the channel length of visual feature blocks and $d$ is the dimension of all LSTM states. $x_t$ is the visual feature refined by estimated facial mask $\alpha_t(k)$. A residual link with adjustable weights $w_{res}$ is included in the facial soft attention module. 

Facial masks are learned with previous hidden state $\pmb{h}_{t-1}$ and visual feature $\pmb{C}_t(k)$. $\pmb{v}$, $\pmb{A}_h$ and $\pmb{A}_c$ are learnable weights:
\begin{equation}
\label{equ:att_mask}
\tilde{\alpha}_t(k)=\pmb{v}\tanh(\pmb{A}_h\pmb{h}_{t-1}+\pmb{A}_c\pmb{C}_t(k)+b)
\end{equation}
\begin{equation}
\alpha_t(k)=\frac{\exp(\tilde{\alpha}_t(k))}{\sum_{j=1}^{H*W} \exp(\tilde{\alpha}_t(j))}
\end{equation}

Different from previous self-attention studies~\cite{xu2015show}, facial attention masks $\alpha_t(k)$ are learned with facial keypoint heatmap supervisions $M_t$ and L2 losses as shown in Eq.~\ref{equ:poseloss}, which provides the clear semantic meaning of facial regions to the learned attention masks.
\begin{equation}
\label{equ:poseloss}
L_{kp}=\sum_{t=1}^T \sum_{k=1}^{H*W}\norm{ \biggl(M_t(k)-\alpha_t(k)\biggr) }^2
\end{equation}
The heatmap $M_t$ is converted from estimated keypoints:
\begin{equation}
\label{equ:heatmap}
M_t=\sum_{j=1}^{J}Conf_t^j * \mathcal{N}(Center_t^j,\sigma)
\end{equation}
where each keypoint $Center_t^j$ is converted into a 2D Gaussian distribution centered at the keypoint. Keypoint estimation confidences $Conf_t^j$ provided by keypoint estimation algorithms are also included in heatmap generation to adjust the weights for Gaussian peaks. Finally, the overlay of keypoint heatmaps is normalized spatially with the softmax function. 

The proposed soft attention fusion method is naturally robust against incomplete keypoints. Moreover, the inaccurate predictions with low confidence scores are depressed by the low keypoint estimation confidence included in the heatmap generation step. Therefore, the proposed approach is robust against both {\it incorrect} and {\it incomplete} estimated keypoints. Furthermore, it improves the method's interpretability by reflecting attended regions.
Correct facial masks can even be generated on cartoon GIFs where no estimated facial keypoints are available. The entire framework is trained end-to-end with intermediate keypoint supervision loss $L_{kp}$ added to main emotion recognition loss in Eq.~\ref{loss_e}:
\begin{equation}
\label{equ:loss}
\mathcal{L}=\mathcal{L}_E+w_{kp}*L_{kp}
\end{equation}
\vspace{-9pt}
\subsection{Hierarchical Segment LSTM Network (HS-LSTM)}
\label{sec:hslstm}
In Section \ref{sec:kavan}, we introduce the base KAVAN with a single LSTM layer. Naive temporal networks tend to forget information in early stages~\cite{bai2018empirical}. This is not desired in GIF emotion recognition, because GIF contains less redundant frames and all frames are indispensable towards correct emotion recognition. Inspired by recent studies~\cite{chung2016hierarchical}, we propose a Hierarchical Segment LSTM module (HS-LSTM) to better model long-term temporal dependencies.

\begin{figure}[t]
\begin{center}
   \centerline{\includegraphics[width=7cm]{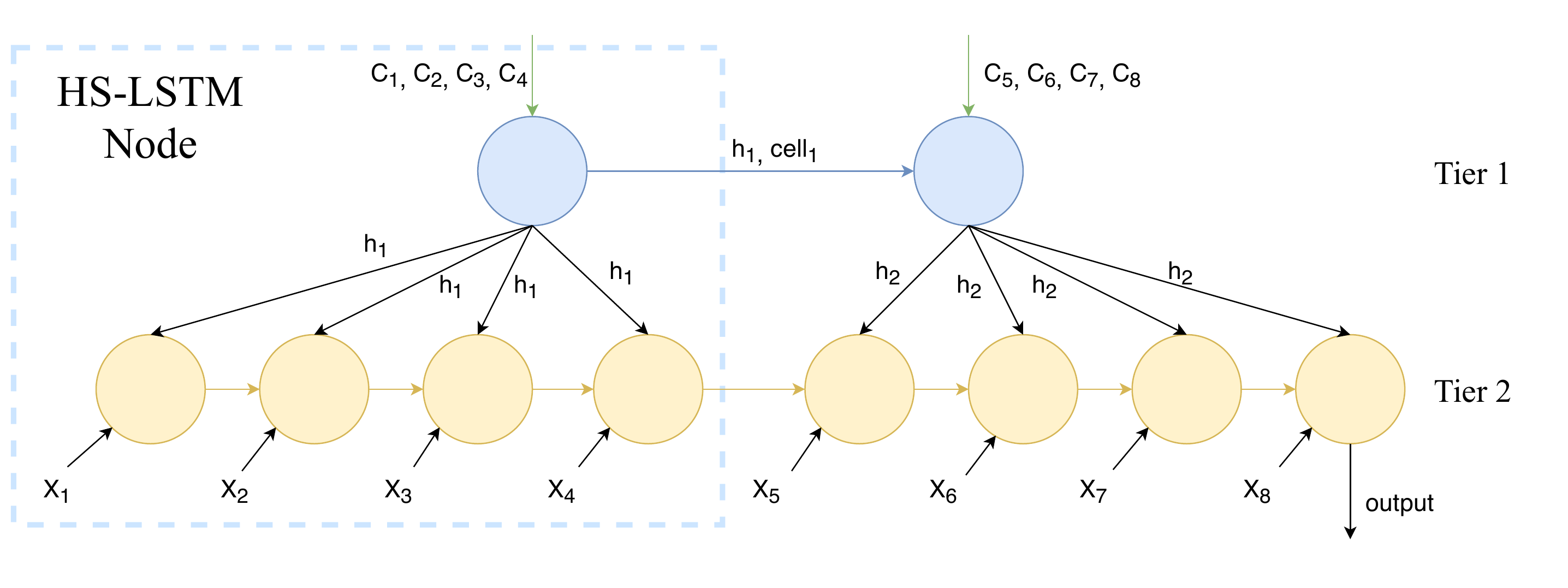}}
\end{center}
\vspace{-0.45in}
	\caption{A two-tier HS-LSTM network structure with two HS-LSTM nodes of size four, where $X$ and $C$ are attended and original visual features.}
\vspace{-0.2in}
\label{fig:hlstm}
\end{figure}
Instead of learning global representations sequentially with LSTM layers, HS-LSTM first generates segment-level representations for each segment in GIFs. The segment-level representations are then propagated through different tiers for global GIF-level representations.  As shown in Fig.~\ref{fig:hlstm}, HS-LSTM contains several tiers of LSTM layers that learns representations from coarse- to fine-grained. The first tier takes the stacked features in a segment to learn a coarse segment representation. 
Nodes in the next tier takes corresponding frame features and the coarse representations learned in the previous tier as input, and learns a refined representation.
The representations learned at different temporal resolutions are then propagated through the HS-LSTM network for a final GIF representation. The number of tiers, HS-LSTM nodes and input frames can be adjusted flexibly based on data statistics. 


Finally, we show the complete KAVAN structure with HS-LSTM module integrated. The keypoint attended visual attention is only conducted in the last tier:
\begin{equation}
\label{equ:all_att}
\tilde{\alpha}_t(k)=\pmb{v}\tanh \left( \pmb{A}_h\pmb{h}_{t-1}+\pmb{A}_H\sum_{l=1}^{Tier-1}\pmb{H}_{l}^i+\pmb{A}_c\pmb{C}_t(k)+b \right )
\end{equation}
where $\pmb{H}_{l}^i$ is the output segment representation in a same segment $i$ at all previous tiers $l$. The input visual feature to the last tier is weight-averaged by the generated attention mask. The inputs to all other tiers remain unchanged.
\vspace{-9pt}
\subsection{Problem Formulation}
\label{sec:Prob_Model}
In this section, we introduce the problem formulation and training objective for GIF emotion recognition. The base task is modeled as the regression of emotion intensities on all labeled emotion classes.
Normalized mean squared error $L_{RG}$ is used for regression, which can avoid over or under prediction~\cite{jou2014predicting} compared to the MSE loss. The normalized mean squared error ($nMSE$) is defined as the mean squared error divided by the variance of the target vector.

\begin{figure}[t]
\begin{center}
   \centerline{\includegraphics[width=7cm]{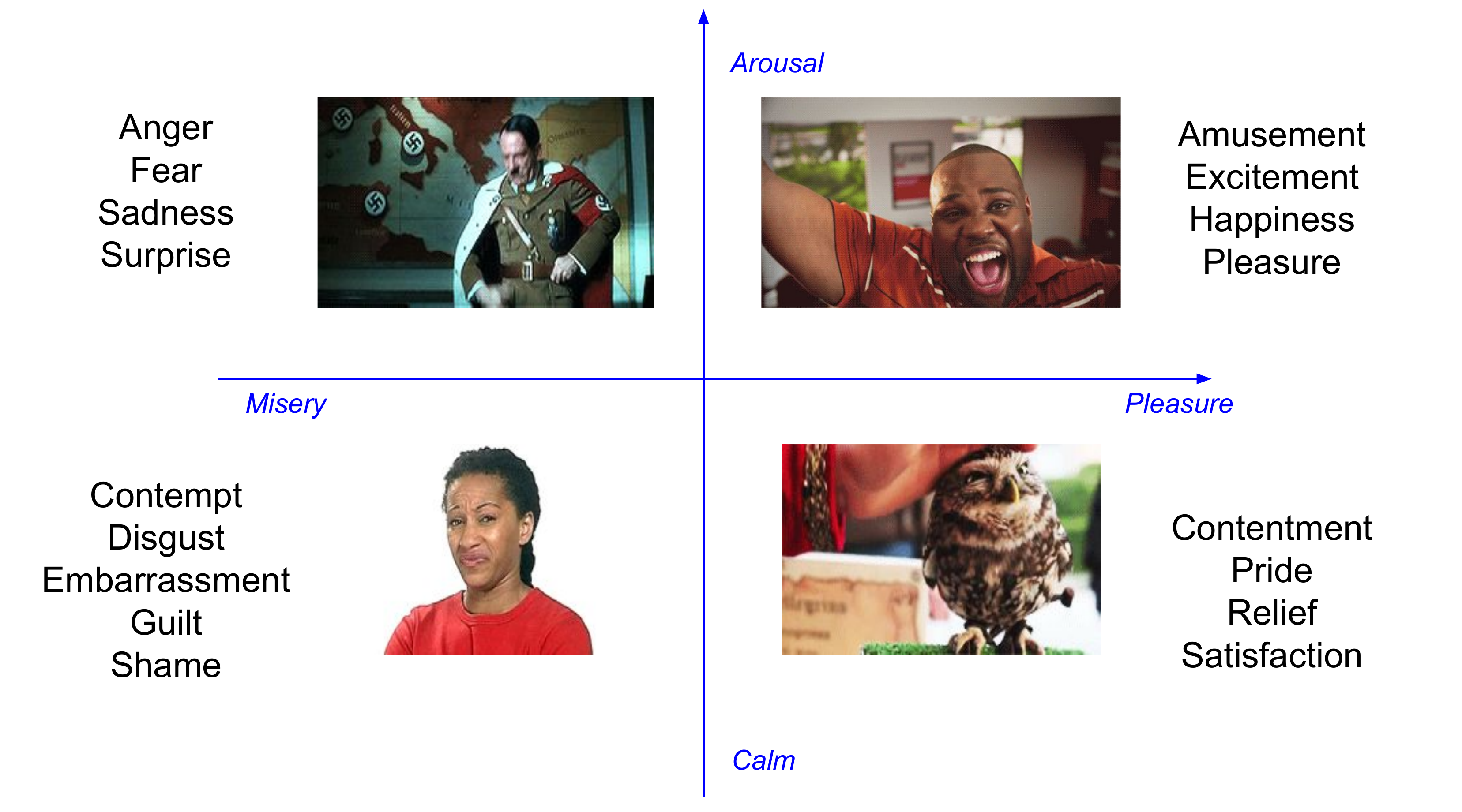}}
\end{center}
\vspace{-0.4in}
	\caption{The four coarse emotion categories summarized from the 17 classes with the circumplex affect model~\cite{russell1980circumplex}.}
\vspace{-0.2in}
\label{fig:class_gif}
\end{figure}
Although intensity score regression is a good formulation for emotion recognition, it becomes increasingly challenging when the number of emotion classes increases. To alleviate this problem and meanwhile achieve a reliable understanding about coarse GIF emotion categories, we divide all labeled emotions into four coarse categories based on the circumflex affect model~\cite{russell1980circumplex,mehrabian1996pleasure}. The circumflex affect model proposes that emotions are distributed in a 2D circular space, where the vertical axis represents `arousal' and the horizontal axis represents `valence'. With the two axes, we divide the emotions into four categories as shown in Fig.~\ref{fig:class_gif}. We conduct a four-class-classification with cross entropy loss $L_C$ alongside the main regression task. 
Introducing the categorical emotion classification task has two advantages. First, predicted coarse emotion labels provide extra prior knowledge to the regression branch and make regression easier.
Second, a reliable classification branch guarantees correct understanding for the coarse emotion type. For example, confusing `Happiness' with `Pleasure' is a smaller error compared to interpreting `Happiness' as a negative emotion.


Finally, we include a ranking loss $L_{RANK}$ to preserve the rank from the strongest emotion to the most unlikely one. We show that predicting the ranking order of emotion intensity scores could also help the regression task. The proposed ranking loss $L_{RANK}$ is consist of the sum of pairwise ranking loss that is designed to penalize the incorrect orders:
\begin{equation}
\label{equ:ranking loss}
L_{RANK}(i)=\left| \left\{ (k,l):\hat{f}_{ik}<\hat{f}_{il}, k<l, k\ in\ K \right\} \right|
\end{equation}
where $\hat{f}$ is the emotion intensity and $K$ is the total number of emotions.
The final loss $\mathcal{L}_E$ for emotion recognition is:
\begin{equation}
\label{loss_e}
\mathcal{L}_E=L_{RG}+w_{C}*L_{C}+w_{RANK}*L_{RANK}
\end{equation}
\section{Experiments}
We first introduce the GIFGIF dataset and facial keypoints pre-processing methods. The proposed framework is then evaluated with both classification and regression metrics. 
\vspace{-9pt}
\subsection{Experiment Settings}
\label{sec:dataset}
The data used in this study is collected from a website built by MIT Media Lab named GIFGIF, and is referred to as `the GIFGIF dataset'. 
Extending from previous definition of eight emotions, 17 emotions as shown in Figure~\ref{fig:class_gif} are labeled to study the more detailed emotions. 
The dataset is labeled by distributed online users. The annotator is presented with a pair of GIFs and is asked whether GIF A, B or neither expresses a specific emotion. At the time of our data collection, we collect 6,119 GIFs with more than 3.2 million user votes. 
The massive user votes are converted to a 17-dimensional soft emotion intensity score with the TrueSkill algorithm~\cite{herbrich2007trueskill}. 
Each output emotion intensity score ranges in $[0,50]$, which is then linearly normalized into $[-1,1]$. 

Besides the appearance feature, estimated facial keypoints are integrated for GIF emotion recognition. 70 facial keypoints are estimated with OpenPose~\cite{cao2017realtime}. We then convert the 70 keypoints in each frame into heatmaps following Eq.~\ref{equ:heatmap}. Each keypoint corresponds to a Gaussian distribution with $\sigma=5$. The weight of each Gaussian is adjusted by the prediction confidence of the keypoints that $Conf_t^j * \mathcal{N}(Center_t^j,\sigma)$.  The keypoints around lips are denser then other facial regions according to the 70-point facial keypoint definition~\cite{cao2017realtime}. Therefore, the weights for the Gaussian distributions around lips is further reduced by $50\%$. The initial heatmap resolution is $64\times 64$ and is later converted to $7\times 7$ after overlaying all Gaussian peaks. 
We randomly split $80\%$ of data for training and the rest for testing.  The averaged performance on five random splits is reported. The processed data will be released~\footnote{https://github.com/zyang-ur/human-centered-GIF}.
\vspace{-9pt}
\subsection{Categorical Emotion Classification}
\label{sec:4_2}
We first evaluate the proposed modules with the coarse emotion category classification task. 
The emotion categories are generated based on the most significant emotion in a GIF. The number of GIFs in each category is $1507/1275/2004/1333$. 
As shown in Table~\ref{table:classification}, we start with a baseline that uses the {\it ResNet-50 + LSTM} structure and only the regression loss $L_{RG}$. A baseline accuracy of $61.47\%$ is achieved. We then evaluate the effectiveness of the proposed soft attention module and HS-LSTM module separately, which are referred to as {\it Soft-Att+LSTM} and {\it ResNet-50 + HS-LSTM}. The soft attention module learns an keypoint guided attention mask defined in Eq.~\ref{equ:att_mask}. The dimension of $\pmb{v}$ is $1\times 32$. $\pmb{A}_h$ and $\pmb{A}_c$ has a parameter size of $32\times L_{hidden}$ and $32\times D_{conv}$, which is both $32\times 2048$ in this study. $L_{hidden}$ is the hidden size of the LSTM and $D_{conv}$ is the channel number of visual features. With purely the soft attention module, the recognition accuracy improves from $61.47\%$ to $63.55\%$. In the HS-LSTM module experiment, we adopt a two-tier structure with two HS-LSTM nodes of size four as shown in Fig.~\ref{fig:hlstm}. 
With purely the HS-LSTM module, the accuracy improves from $61.47\%$ to $62.55\%$.
When combining the soft attention module with HS-LSTM, the KAVAN framework achieves an accuracy of $65.95\%$, which is better than both separate modules. Furthermore, by incorporating  multi-task learning with the loss proposed in Eq.~\ref{equ:loss}, an extra $2.32\%$ improvements is obtained and the accuracy reaches $68.27\%$. This proves the effectiveness of the proposed MTL setting on the classification task. 

\vspace{-9pt}
\subsection{Multi-task Emotion Regression}
\label{sec:reg}

We then show the effectiveness of the proposed modules and the multi-task learning setting with regression metrics. As shown in Table~\ref{table:regression}, the baseline model {\it ResNet-50 + LSTM} achieves an {\it nMSE} of $1.0675$. 
With the same parameters in Section \ref{sec:4_2}, soft attention module {\it Soft-Att + LSTM} achieves an {\it nMSE} of $1.0100$, which is significantly better than the baseline. HS-LSTM module {\it ResNet-50 + HS-LSTM} along also outperforms the baseline LSTM by reaching an {\it nMSE} of $1.0277$. Finally, we evaluate the full framework with both the soft attention module and HS-LSTM adopted. The method reaches an {\it nMSE} of $0.9987$. 
\begin{table}[t]
\begin{center}
\caption{The coarse emotion category classification accuracy.}
\begin{tabular}{ c c }
	\hline
    Methods & Accuracy $\%$\\
	\hline
    ResNet-50 + LSTM & $61.47\%$\\ 
    Soft-Att + LSTM & $62.55\%$\\ 
    ResNet-50 + HS-LSTM & $63.55\%$\\ 
    Soft-Att + HS-LSTM & $65.95\%$\\
	MTL Soft-Att + HS-LSTM & $\bm{68.27}\%$\\ 
	\hline  
\end{tabular}
\end{center}
\vspace{-0.3in}
\label{table:classification}
\end{table}
\begin{table}[t]
\vspace{0.05in}
\begin{center}
\caption{The predicted Normalized Mean Squared Error (nMSE) on the GIFGIF dataset.}
\vspace{0.05in}
\begin{tabular}{ c c c}
    \hline
    Methods & Loss \hphantom & nMSE\\
    \hline
    Color Histogram~\cite{jou2014predicting} & nMSE & $1.2623 \pm 0.0615$\\
    Face Expression~\cite{jou2014predicting} & nMSE & $1.0000 \pm 0.0064$\\
	\hline
    ResNet-50 + LSTM & nMSE & $1.0675 \pm 0.0305$\\
    Soft-Att + LSTM & nMSE & $1.0100 \pm 0.0085$\\
    ResNet-50 + HS-LSTM & nMSE & $1.0277 \pm 0.0147$\\
    Soft-Att + HS-LSTM & nMSE & $0.9987 \pm 0.0033$\\	
    Soft-Att + HS-LSTM & MTL & $\pmb{0.9863 \pm 0.0084}$\\	
    \hline  
\end{tabular}
\end{center}
\vspace{-0.3in}
\label{table:regression}
\end{table}
Furthermore, we fuse the regression framework with the classification branch by conducting multi-task learning.
A weighted sum of the {\it nMSE} loss, the {\it CrossEntropy} loss and the {\it ranking} loss is adopted to train the framework. An extra improvement is obtained and the {\it nMSE} reaches $0.9863$. 

We then compare our results to other state-of-the-art on MIT GIFGIF. Because the GIFGIF dataset keeps growing, the version we collect with 6,119 GIFs is larger than the one in previous study~\cite{jou2014predicting} with 3,858 GIFs. Therefore, a direct comparison is unfair as the task becomes more challenging with more ambiguous GIFs included. 
Based on our re-implementation as shown in Table~\ref{table:regression}, the {\it Face Expression + Ordinary Least Squares Regression} approach~\cite{jou2014predicting} works the best and achieves a {\it nMSE} of $1.0000$. OpenCV's haar feature-based cascade classifiers are used for face detection. CNN+SVM facial expression features~\cite{tang2013deep,jou2014predicting} pretrained on a facial emotion dataset~\cite{tang2013deep} are extracted on the largest detected face. Our proposed KAVAN framework achieves a {\it nMSE} of $0.9863$, which is better than the best re-implemented statr-of-the-art of $1.0000$. 
\vspace{-9pt}
\subsection{Qualitative Results}
\begin{figure}[t]
\begin{center}
   \centerline{\includegraphics[width=8.6cm]{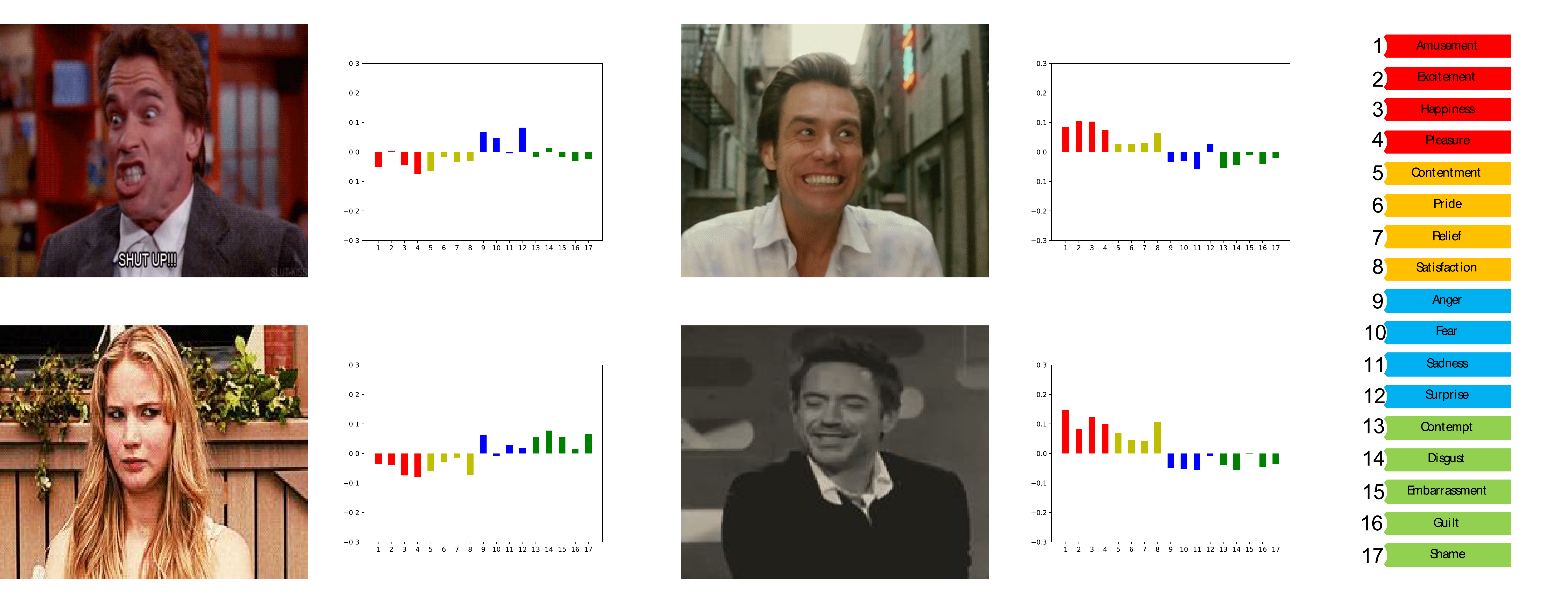}}
\end{center}
\vspace{-0.4in}
	\caption{The visualization of GIF emotion recognition results.}
\vspace{-0.1in}
\label{fig:example}
\end{figure}

As shown in Fig.~\ref{fig:example}, good qualitative results are observed. For example, the upper-left GIF in Fig.~\ref{fig:example} belongs to category `Misery Arousal' that is represented in blue, and is predicted correctly. Fittingly, the predicted emotion intensity of `anger', `fear' and `supervise' are the highest.
\begin{figure}[t]
\begin{center}
   \centerline{\includegraphics[width=9cm]{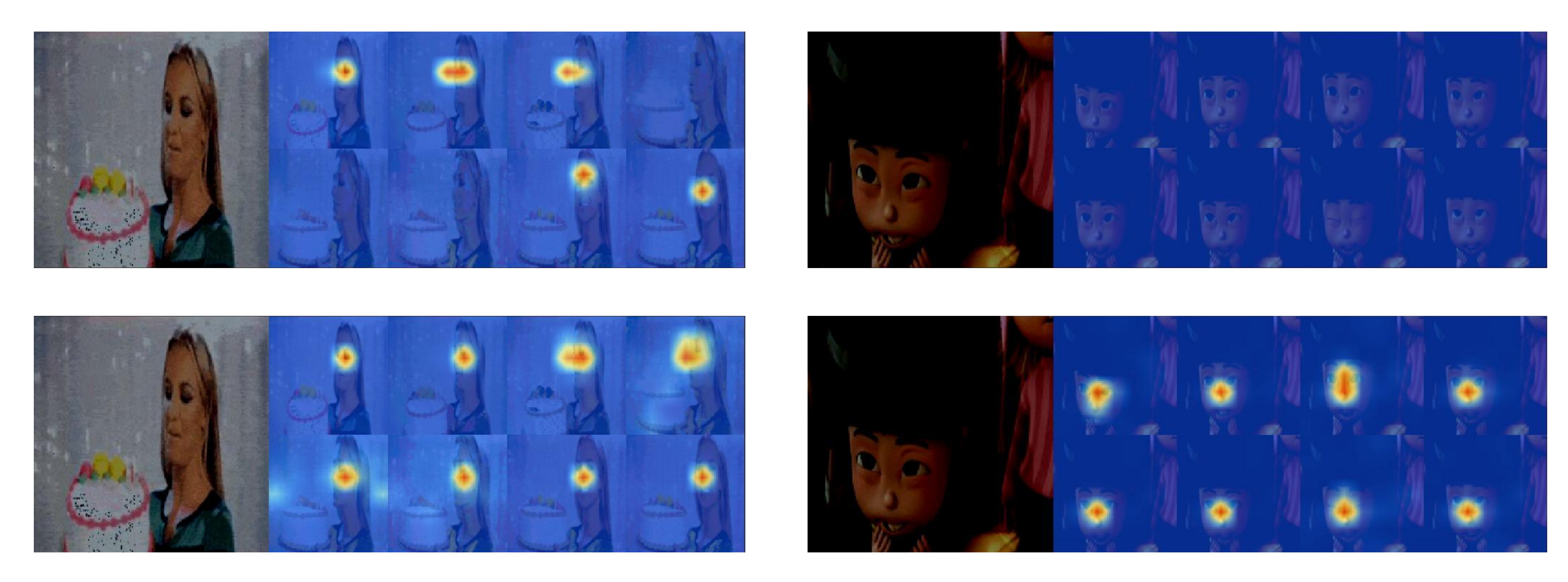}}
\end{center}
\vspace{-0.4in}
	\caption{The visualization of predicted facial attention masks. Upper figures show supervision heatmaps and lower figures visualize the estimated facial masks predicted by KAVAN.}
\vspace{-0.1in}
\label{fig:attention}
\end{figure}
Furthermore, ideal results on facial region estimation is also observed, as shown in Fig.~\ref{fig:attention}.
The larger unmasked image on the left of each sub-figure is the first sampled input frame, and the remaining eight smaller images are the overlay of input frames and facial keypoint heatmaps. 
The upper two sub-figures in Fig.~\ref{fig:attention} visualize the supervision heatmaps generated with estimated facial keypoints, which may be inaccurate or incomplete. As shown in the lower two sub-figures in Fig.~\ref{fig:attention}, the attention masks predicted by KAVAN accurately focus on correct facial regions even when no original keypoint annotations are available, such as in cartoon GIFs.
Experiments show that the proposed approach well utilizes the keypoints information and is robust against missing or inaccurate annotations. Furthermore, the predicted facial region masks improve the framework's interpretability.
\section{Conclusion}
Motivated by GIF's unique properties, we focus on human-centered GIF emotion recognition and propose a Keypoint Attended Visual Attention Network (KAVAN). In the facial  attention module, we learn facial region masks with estimated facial keypoints to guide the GIF frame representation extraction. In the temporal module, we propose a novel Hierarchical Segment LSTM (HS-LSTM) structure to better represent the temporal evolution and learn better global representations. Experiments on the GIFGIF dataset validate the effectiveness of the proposed framework.\\


\noindent\footnotesize\textbf{Acknowledgement}. This work is partially supported by NSF awards \#1704309, \#1722847, and \#1813709.

\bibliographystyle{IEEEbib}
{\footnotesize
\bibliography{sample-bibliography}}

\begin{thebibliography}{10}

\bibitem{bakhshi2016fast}
Saeideh Bakhshi, David~A Shamma, Lyndon Kennedy, Yale Song, Paloma de~Juan, and
  Joseph'Jofish' Kaye,
\newblock ``Fast, cheap, and good: Why animated gifs engage us,''
\newblock in {\em CHI}. ACM, 2016, pp. 575--586.

\bibitem{chen2017gifgif+}
Weixuan Chen, Ognjen~Oggi Rudovic, and Rosalind~W Picard,
\newblock ``Gifgif+: Collecting emotional animated gifs with clustered
  multi-task learning,''
\newblock in {\em ACII}. IEEE, 2017, pp. 410--417.

\bibitem{jou2014predicting}
Brendan Jou, Subhabrata Bhattacharya, and Shih-Fu Chang,
\newblock ``Predicting viewer perceived emotions in animated gifs,''
\newblock in {\em ACM MM}. ACM, 2014, pp. 213--216.

\bibitem{chen2016predicting}
Weixuan Chen and Rosalind~W Picard,
\newblock ``Predicting perceived emotions in animated gifs with 3d
  convolutional neural networks,''
\newblock in {\em ISM}. IEEE, 2016, pp. 367--368.

\bibitem{tang2013deep}
Yichuan Tang,
\newblock ``Deep learning using linear support vector machines,''
\newblock {\em arXiv preprint arXiv:1306.0239}, 2013.

\bibitem{jhuang2013towards}
Hueihan Jhuang, Juergen Gall, Silvia Zuffi, Cordelia Schmid, and Michael~J
  Black,
\newblock ``Towards understanding action recognition,''
\newblock in {\em ICCV}. IEEE, 2013, pp. 3192--3199.

\bibitem{cheron2015p}
Guilhem Ch{\'e}ron, Ivan Laptev, and Cordelia Schmid,
\newblock ``P-cnn: Pose-based cnn features for action recognition,''
\newblock in {\em ICCV}, 2015, pp. 3218--3226.

\bibitem{gygli2016video2gif}
Michael Gygli, Yale Song, and Liangliang Cao,
\newblock ``Video2gif: Automatic generation of animated gifs from video,''
\newblock in {\em CVPR}, 2016.

\bibitem{zhao2014exploring}
Sicheng Zhao, Yue Gao, Xiaolei Jiang, Hongxun Yao, Tat-Seng Chua, and Xiaoshuai
  Sun,
\newblock ``Exploring principles-of-art features for image emotion
  recognition,''
\newblock in {\em ACM MM}. ACM, 2014, pp. 47--56.

\bibitem{zhao2017continuous}
Sicheng Zhao, Hongxun Yao, Yue Gao, Rongrong Ji, and Guiguang Ding,
\newblock ``Continuous probability distribution prediction of image emotions
  via multitask shared sparse regression,''
\newblock {\em IEEE Transactions on Multimedia}, vol. 19, no. 3, pp. 632--645,
  2017.

\bibitem{you2016building}
Quanzeng You, Jiebo Luo, Hailin Jin, and Jianchao Yang,
\newblock ``Building a large scale dataset for image emotion recognition: The
  fine print and the benchmark.,''
\newblock in {\em AAAI}, 2016, pp. 308--314.

\bibitem{rao2016learning}
Tianrong Rao, Min Xu, and Dong Xu,
\newblock ``Learning multi-level deep representations for image emotion
  classification,''
\newblock {\em arXiv preprint arXiv:1611.07145}, 2016.

\bibitem{zhao2017learning}
Sicheng Zhao, Guiguang Ding, Yue Gao, and Jungong Han,
\newblock ``Learning visual emotion distributions via multi-modal features
  fusion,''
\newblock in {\em ACM MM}. ACM, 2017, pp. 369--377.

\bibitem{han2017hard}
Jing Han, Zixing Zhang, Maximilian Schmitt, Maja Pantic, and Bj{\"o}rn
  Schuller,
\newblock ``From hard to soft: Towards more human-like emotion recognition by
  modelling the perception uncertainty,''
\newblock in {\em ACM MM}. ACM, 2017, pp. 890--897.

\bibitem{bakhshi2014faces}
Saeideh Bakhshi, David~A Shamma, and Eric Gilbert,
\newblock ``Faces engage us: Photos with faces attract more likes and comments
  on instagram,''
\newblock in {\em CHI}. ACM, 2014, pp. 965--974.

\bibitem{wang2016temporal}
Limin Wang, Yuanjun Xiong, Zhe Wang, Yu~Qiao, Dahua Lin, Xiaoou Tang, and Luc
  Van~Gool,
\newblock ``Temporal segment networks: Towards good practices for deep action
  recognition,''
\newblock in {\em ECCV}. Springer, 2016, pp. 20--36.

\bibitem{he2016deep}
Kaiming He, Xiangyu Zhang, Shaoqing Ren, and Jian Sun,
\newblock ``Deep residual learning for image recognition,''
\newblock in {\em CVPR}, 2016, pp. 770--778.

\bibitem{xu2015show}
Kelvin Xu, Jimmy Ba, Ryan Kiros, Kyunghyun Cho, Aaron Courville, Ruslan
  Salakhudinov, Rich Zemel, and Yoshua Bengio,
\newblock ``Show, attend and tell: Neural image caption generation with visual
  attention,''
\newblock in {\em ICML}, 2015, pp. 2048--2057.

\bibitem{bai2018empirical}
Shaojie Bai, J~Zico Kolter, and Vladlen Koltun,
\newblock ``An empirical evaluation of generic convolutional and recurrent
  networks for sequence modeling,''
\newblock {\em arXiv preprint arXiv:1803.01271}, 2018.

\bibitem{chung2016hierarchical}
Junyoung Chung, Sungjin Ahn, and Yoshua Bengio,
\newblock ``Hierarchical multiscale recurrent neural networks,''
\newblock in {\em ICLR}, 2017.

\bibitem{russell1980circumplex}
James~A Russell,
\newblock ``A circumplex model of affect.,''
\newblock {\em Journal of personality and social psychology}, vol. 39, no. 6,
  pp. 1161, 1980.

\bibitem{mehrabian1996pleasure}
Albert Mehrabian,
\newblock ``Pleasure-arousal-dominance: A general framework for describing and
  measuring individual differences in temperament,''
\newblock {\em Current Psychology}, vol. 14, no. 4, pp. 261--292, 1996.

\bibitem{herbrich2007trueskill}
Ralf Herbrich, Tom Minka, and Thore Graepel,
\newblock ``Trueskill™: a bayesian skill rating system,''
\newblock in {\em NIPS}, 2007, pp. 569--576.

\bibitem{cao2017realtime}
Zhe Cao, Tomas Simon, Shih-En Wei, and Yaser Sheikh,
\newblock ``Realtime multi-person 2d pose estimation using part affinity
  fields,''
\newblock in {\em CVPR}, 2017.

\end{thebibliography}

\end{document}